\documentclass[a4paper,fleqn]{cas-dc}

\usepackage{placeins}

\usepackage[numbers]{natbib}
\usepackage[utf8]{inputenc}
\usepackage{epstopdf}
\usepackage[figuresright]{rotating}
\usepackage{threeparttable}
\usepackage{color}
\usepackage{makecell}
\usepackage{mathrsfs}
\usepackage{amsfonts}
\usepackage{algorithm}
\usepackage{algpseudocode}
\usepackage{url}
\usepackage{amsmath}
\usepackage{booktabs}
\usepackage{multirow}
\usepackage{tikz}
\usetikzlibrary{fit,shapes}

\usepackage{graphicx}
\usepackage{epstopdf}  
\usepackage{capt-of}

\cortext[cor1]{Corresponding author: Yuan-Gen Wang (wangyg@gzhu.edu.cn)}
\cortext[cor2]{Corresponding author: Ruping Wang (dxcyjc@163.com)}

\begin{document}
\shorttitle{L-VARC for Visual Reasoning}
\shortauthors{Ye et~al.}

\title [mode = title]{Language-Guided Abstraction for Visual Reasoning}                      


\author[1]{Xu-Jing Ye}
\author[1]{Yuan-Gen Wang}   
\cormark[1]          
\ead{wangyg@gzhu.edu.cn}

\author[2]{Ruping Wang}
\cormark[1]
\ead{dxcyjc@163.com}

\affiliation[1]{organization={School of Artificial Intelligence, Guangzhou University},
               city={Guangzhou},
               postcode={510006},
               country={China}}

\affiliation[2]{organization={Traditional Chinese Medicine Hospital of Zengcheng District},
               city={Guangzhou},
               postcode={511399},
               country={China}}

\begin{abstract}
The Abstraction and Reasoning Corpus (ARC) is viewed as a critical avenue to Artificial General Intelligence (AGI), as it enables models to learn abstract transformation rules from few-shot examples and then generalize to new tasks. However, prevalent ARC methodology is either pure language or vision-only (i.e., VARC). The former depends heavily on LLMs, consuming billions of parameters. The latter often struggles to capture high-level semantics, leading to overfitting on pixel-level patterns. To bridge this gap, we propose \textbf{L-VARC}, a novel framework that enhances visual reasoning via a language-guided \textit{Learning Using Privileged Information (LUPI)} branch. Specifically, we design a \textbf{Semantic Compression Module} by feeding a unified, task-agnostic prompt into DeepSeek-V3. In this way, the raw LARC (a crowd-sourced language description dataset) can be substantially refined and structured, fitting with the context length constraint of standard text encoders (e.g., CLIP). Moreover, we design a \textbf{Cross-Attention Projector} to align visual features with semantic embeddings, aiming to guide the training of the ARC model. Notably, the LUPI branch is taken in the training process and will be discarded during inference, thereby yielding a lightweight model with a mere 18 million parameters. Extensive experiments demonstrate that our L-VARC effectively leverages linguistic priors to boost visual reasoning and outperforms state-of-the-art. Ablation studies further confirm the contribution of the two new designs towards the L-VARC framework. The code is available at https://github.com/GZHU-DVL/L-VARC.
\end{abstract}


\maketitle

\begin{center}
    \includegraphics[width=\columnwidth, trim=10 0 10 30, clip]{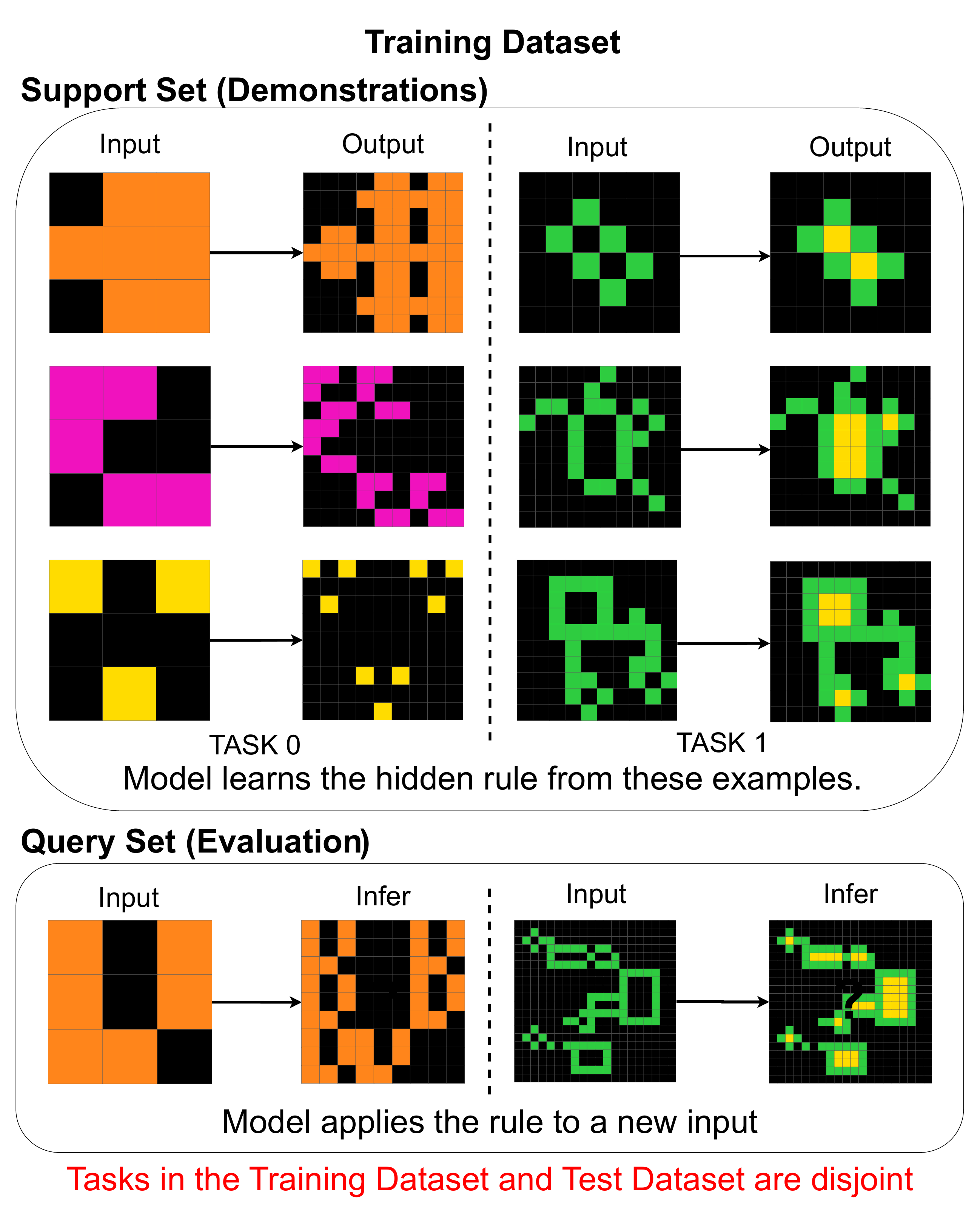}
    \vspace{-2.5em} 
    \captionof{figure}{Sample tasks from the ARC training set. \textbf{Task 0} involves a rule of \textit{positional extension}, while \textbf{Task 1} requires \textit{filling enclosed regions}. The model must infer these hidden rules from the Support Set to solve the Query Set.}
    \label{fig:task}
\end{center}

\section{Introduction}
Learning abstract concepts from few-shot demonstrations is a core challenge in machine intelligence. The Abstraction and Reasoning Corpus (ARC) \cite{chollet2019measure} benchmarks such capability through puzzle-like tasks governed by hidden transformation rules, as illustrated in Fig. \ref{fig:task}. While recent research has framed ARC primarily as a program synthesis problem leveraging Large Language Models (LLMs) \cite{chollet2024arc,wang2023hypothesis,ellis2021dreamcoder}, these approaches typically consume billions of parameters and struggle to exploit the purely visual intuition required for many tasks. In contrast, Vision ARC (VARC) framework \cite{hu2025arc} successfully reformulated it as a pure image-to-image translation task. By leveraging standard Vision Transformers (ViT) \cite{dosovitskiy2020image} and Test-Time Training (TTT) \cite{bottou1992local,sun2020test}, VARC demonstrated that visual priors (e.g., locality, translation, and scaling invariance) are crucial for solving physical-like puzzles. However, a purely vision-centric approach faces an inherent limitation. Without high-level semantic guidance, models tend to overfit to spurious pixel-level correlations rather than learning the abstract concepts (e.g., ``gravity", ``intersection") required to generalize to the disjoint test set (see Fig.~\ref{fig:test}).

The VARC framework has recently demonstrated that a lightweight ViT with TTT can achieve competitive performance on ARC-1 \cite{hu2025arc}. However, its purely vision-centric nature leads to two observable shortcomings when faced with unseen test tasks. First, as shown in our qualitative analysis (Fig.~\ref{fig:qualitative}, Case A), VARC struggles to remove distractors—such as new colors appearing only in the test set—and often fails to transfer the correct abstract rule (e.g., ``crop the green object''). In contrast, our L-VARC successfully applies the same rule to the test input, converging to the correct solution with significantly higher voting confidence (48 votes vs.~6 votes). Second, on tasks requiring dense pixel manipulation (e.g., filling enclosed regions, Case B), VARC produces incomplete outputs with missing spots, whereas our L-VARC completely fills all target regions. These observations suggest that without high-level semantic guidance, visual models tend to overfit to spurious correlations and lack the precision needed for fine-grained transformations. Recent work \cite{li2025tackling} further points out that standard ViTs may have inherent representational deficiencies for abstract visual reasoning, which pure vision-based methods struggle to overcome.

\begin{center}
   \includegraphics[width=\columnwidth, trim=0 0 0 0, clip]{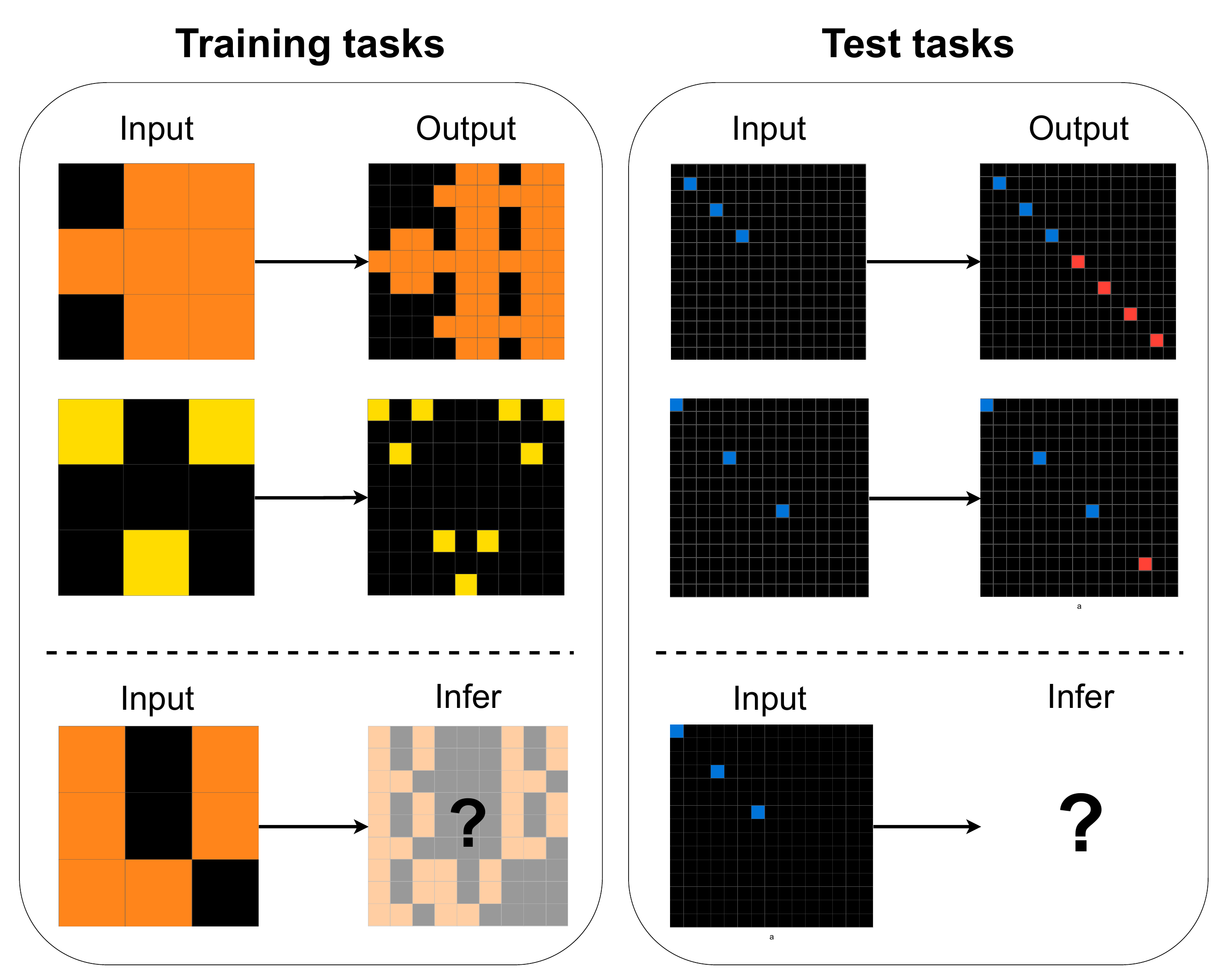}
   \vspace{-2em}
   \captionof{figure}{\textbf{Visualization of the generalization gap.} Training tasks typically involve specific core knowledge priors (e.g., object persistence), while test tasks require combining these priors into novel abstract rules. This strict separation necessitates a model capable of meta-learning rather than simple pattern recognition.}
   \label{fig:test}
\end{center}

These limitations are further reflected in quantitative metrics. On the harder ARC-2 benchmark, which is designed to combine core knowledge priors in novel ways, VARC's pass@1 accuracy is only 6.25\% (Table~\ref{tab:main_results}). By injecting linguistic descriptions as privileged information, our L-VARC improves the pass@1 accuracy to 6.67\%. While a +0.42\% gain seems modest, most LLMs struggle on ARC-2 (see Table~\ref{tab:llm_comparison}). Moreover, our per-category analysis (Sec.~\ref{sec:percat}) shows that the improvement is concentrated on tasks requiring abstract rule induction (\textit{Transformation} and \textit{Filling}), confirming that language guidance specifically benefits generalization to unseen concepts.

To bridge the gap between pixel-level processing and abstract reasoning, we propose L-VARC, a language-guided VARC, by resorting to the hypothesis that vision excels at pattern abstraction while language specializes in rule formulation \cite{zhang2025vislang,jia2025duet}. Concretely, we cast ARC as a Learning Using Privileged Information (LUPI) problem \cite{vapnik2009new}, where the LARC dataset \cite{acquaviva2022communicating} provides natural language descriptions that serve as privileged information during training. Recognizing that raw descriptions are often noisy or verbose, we design a Semantic Compression Module (SCM) by using DeepSeek-V3 \cite{liu2024deepseek} to distill these descriptions into concise and rule-centric statements. These refined texts are then encoded by a frozen CLIP text encoder \cite{radford2021learning} to obtain compact semantic embeddings. To effectively align visual features with these embeddings, we design a Cross-Attention Projector (CAP) on top of the ViT encoder. Instead of simple Global Average
Pooling (GAP), our CAP uses a learnable query to dynamically attend to semantically relevant visual patches and aligns them with the CLIP text embeddings \cite{zhang2023cross}. This process forces the visual encoder to capture concept-level features that are consistent with linguistic descriptions. Note that this language-guided LUPI branch can be discarded during inference, without increasing any model size. It is worth emphasizing that our work follows the LUPI paradigm \cite{vapnik2009new, pechyony2010theory}, where additional information in the form of language descriptions is available only during training. Unlike standard multi‑modal learning, LUPI does not require the privileged information at test time, which perfectly suits ARC: linguistic descriptions of hidden rules are available for training tasks (via LARC) but not for unseen test tasks. This makes our method distinguish from pure vision or pure language approaches. The major contributions of this paper can be summarized as:
\begin{itemize}
    \item We propose L-VARC, a language-guided VARC framework that enhances VARC with distilled linguistic knowledge from DeepSeek and CLIP, resulting in the same model size as VARC.
    \item We design a semantic compression module (SCM) to condense the raw LARC descriptions into precise rule statements, yielding robust semantic embeddings.
    \item We design a cross-attention projector (CAP) to align semantic embedding and visual representation, effectively sharpening visual features for abstract reasoning.
    \item Extensive experiments on ARC‑1 and ARC‑2 demonstrate an improved PASS@1 over the baseline, along with a validation accuracy increase from 76.7\% to 78.1\%, highlighting the benefit of language‑guided training.
\end{itemize}  

\FloatBarrier
\section{Proposed Method}
\label{sec:method}
\begin{figure*}[]
\centering
\includegraphics[width=6.6in,height=3in,keepaspectratio]
{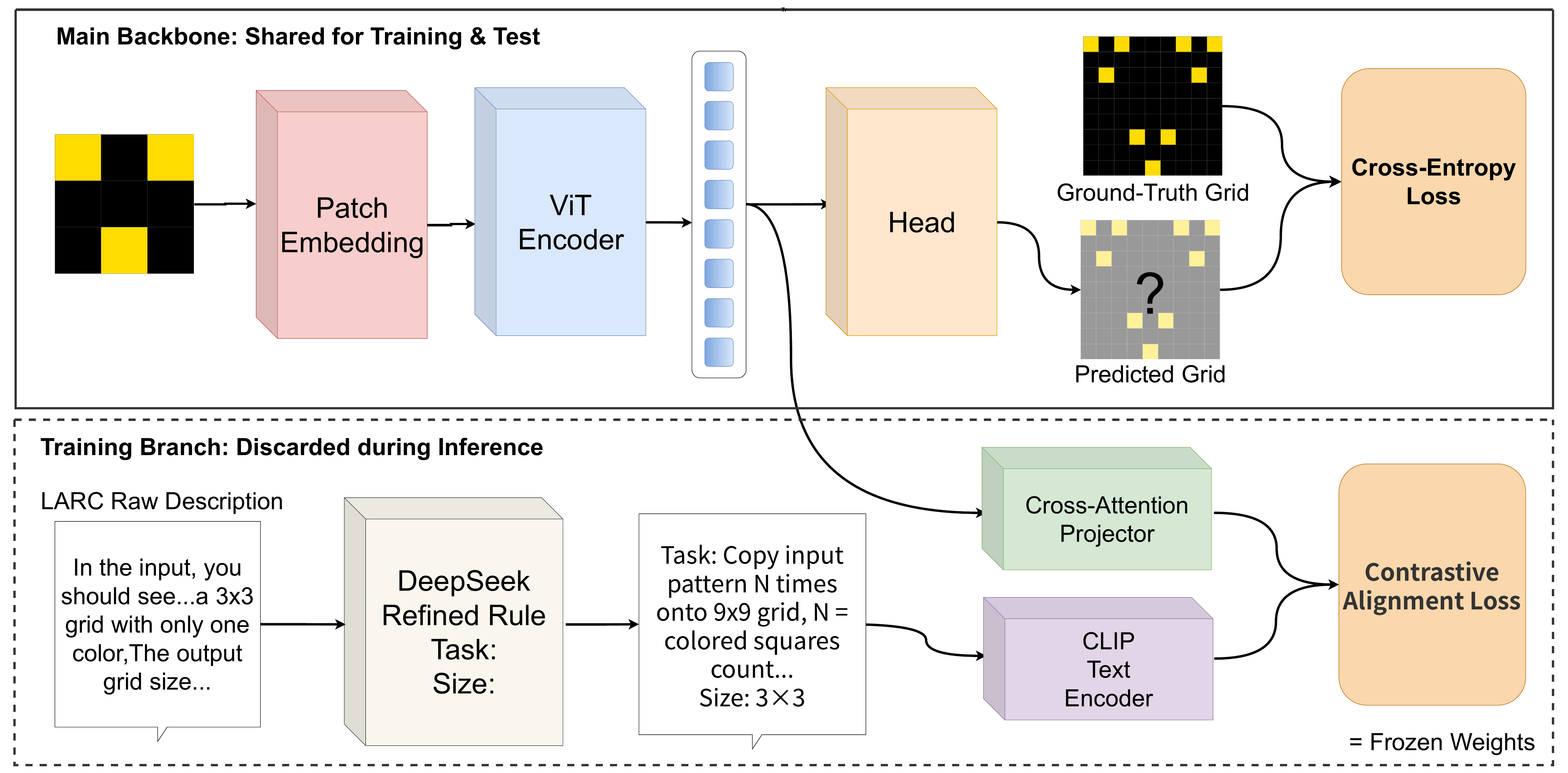}
\vspace{-0.5em}
\caption{Overview of the proposed L-VARC framework. The architecture is organized into two branches. \textbf{Top (Main Backbone):} A shared Vision Transformer (ViT) encodes the input grid and predicts the output grid, optimized by the task reconstruction loss. This branch is used for both training and inference. \textbf{Bottom (Training Branch):} A language-guided training branch injects semantic knowledge. Raw descriptions are refined by DeepSeek-V3 and encoded by a frozen CLIP encoder. Simultaneously, a Cross-Attention Projector extracts concept-specific visual features. These two modalities are aligned via a contrastive loss. Note that the entire bottom branch is discarded during inference.}
\label{fig:framework}
\vspace{-0.5cm}
\end{figure*}

In this section, we detailedly illustrate L-VARC, a new framework designed to enhance few-shot abstract reasoning by aligning visual representations with linguistic concepts. As illustrated in Fig.~\ref{fig:framework}, our framework consists of a primary visual \textbf{Main Backbone} and an auxiliary language-guided \textbf{Training Branch}.

\subsection{Problem Formulation}
The ARC task is defined as a few-shot generalization problem. A task $\mathcal{T}$ consists of a support set $S = \{(x_i^s, y_i^s)\}_{i=1}^K$ containing input-output pairs that exemplify a transformation rule, and a query set $Q = \{(x_j^q, y_j^q)\}_{j=1}^M$ where only the input is given. The goal is to predict the query output $\hat{y}_j^q$ based on $x_j^q$ and the support set $S$.
Standard vision-based approaches (VARCs) encode the grid $x$ into visual features and decode them into a predicted grid. However, relying solely on pixel-level supervision often leads to overfitting on spurious spatial correlations. To address this, we propose integrating privileged linguistic information during training.

\subsection{Language-Guided Training}
Our core innovation is the injection of semantic knowledge into the visual latent space via a dedicated alignment branch which includes Semantic Compression Module and Cross-Attention Projector. Interestingly, this branch operates only during training and can be removed at the inference stage.

\noindent
\textbf{Semantic Compression Module.} The original LARC dataset provides crowd-sourced text descriptions for ARC tasks. However, raw human descriptions are often noisy or verbose. To obtain high-quality semantic targets, we introduce a two-step offline refinement strategy:
\begin{itemize}
    \item Refinement via DeepSeek-V3: We utilize the DeepSeek-V3 LLM \cite{liu2024deepseek} to summarize raw descriptions into concise and rule-oriented statements (e.g., \textit{``Fill the enclosed area with blue"}), resorting to the reasoning capabilities of LLMs \cite{wei2022chain}.
    \item Encoding via CLIP: The refined text $T$ is encoded by a pre-trained, frozen CLIP text encoder \cite{radford2021learning}. Alternatives like BLIP \cite{li2022blip} could also be considered. We obtain the global text embedding $z_{text} \in \mathbb{R}^{D}$, which serves as the semantic anchor. Here, $D=512$ is the output dimension of CLIP text encoder.
\end{itemize}

\noindent
\textbf{Cross-Attention Projector.} To align the visual features with $z_{text}$, simple Global Average Pooling (GAP) of the ViT output is insufficient, as the transformation rule often involves specific objects or regions (e.g., specific colors or shapes). To address this issue, we design a new \textbf{Cross-Attention Projector} to dynamically extract semantically relevant features. Specifically, let $H_{vis} \in \mathbb{R}^{L \times D}$ denote the sequence of visual features output by the ViT encoder, where $L$ is the sequence length. 
We define a learnable query token $Q_{lat} \in \mathbb{R}^{1 \times D}$. The visual features serve as both keys and values via linear projections: $K = H_{vis}W_K$ and $V = H_{vis}W_V$. The projector computes the attention as:
\begin{equation}
    \begin{split}
        z'_{vis} &= \text{Attention}(Q_{lat}, K, V) \\
        &= \text{softmax}\left(\frac{Q_{lat} K^T}{\sqrt{D}}\right) V,
    \end{split}
\end{equation}
where $W_K, W_V \in \mathbb{R}^{D \times D}$ are learnable projection matrices.
The output $z'_{vis}$ is then projected via an MLP to match the dimension of the text embedding, yielding the final visual representation $z_{vis} \in \mathbb{R}^{D}$. This mechanism allows the model to ``focus'' on the visual patches with the linguistic rule.

\subsection{Training Objectives}
The model is optimized using a composite objective function that jointly supervises visual reconstruction and cross-modal alignment.

\noindent
\textbf{Task Reconstruction Loss.} The main backbone performs pixel-wise classification. Each grid pixel takes one of $C=10$ color indices. Let $\hat{y}_{p,c}$ be the predicted probability for color $c$ at pixel $p$, and $y_{p,c}\in\{0,1\}$ the ground-truth one-hot label. The loss is the categorical cross-entropy averaged over all $N_{\text{pix}}$ pixels:
\begin{equation}
    \mathcal{L}_{\text{task}} = -\frac{1}{N_{\text{pix}}} \sum_{p=1}^{N_{\text{pix}}} \sum_{c=1}^{C} y_{p,c} \log \hat{y}_{p,c}.
\label{eq:task}
\end{equation}

\noindent
\textbf{Contrastive Alignment Loss.} To align the visual embedding $z_{\text{vis}}$ (from CAP) with the text embedding $z_{\text{text}}$ (from frozen CLIP), we use a symmetric InfoNCE loss \cite{oord2018representation}. For a batch of $B$ tasks, $(z_{\text{vis}}^{(i)}, z_{\text{text}}^{(i)})$ is the positive pair. The loss is:
\begin{align}
    \mathcal{L}_{\text{v2t}} &= -\frac{1}{B}\sum_{i=1}^{B}\log\frac{\exp(\operatorname{sim}(z_{\text{vis}}^{(i)},z_{\text{text}}^{(i)})/\tau)}{\sum_{j}\exp(\operatorname{sim}(z_{\text{vis}}^{(i)},z_{\text{text}}^{(j)})/\tau)},\\
    \mathcal{L}_{\text{t2v}} &= -\frac{1}{B}\sum_{i=1}^{B}\log\frac{\exp(\operatorname{sim}(z_{\text{text}}^{(i)},z_{\text{vis}}^{(i)})/\tau)}{\sum_{j}\exp(\operatorname{sim}(z_{\text{text}}^{(i)},z_{\text{vis}}^{(j)})/\tau)},
\end{align}
and $\mathcal{L}_{\text{align}} = \frac{1}{2}(\mathcal{L}_{\text{v2t}}+\mathcal{L}_{\text{t2v}})$. Here $\operatorname{sim}(\cdot,\cdot)$ is cosine similarity, and $\tau=0.07$ is the temperature. This loss pulls the visual representation of a task toward its language description while pushing it away from descriptions of other tasks.

\noindent
\textbf{Intuition.} The alignment loss acts as a semantic regularizer, encouraging the visual encoder to capture concept-level features consistent with linguistic rules. Without it, the model tends to overfit to pixel-level spurious correlations. Empirically, adding $\mathcal{L}_{\text{align}}$ improves PASS@1 by $+0.87\%$ on ARC-1 and $+0.42\%$ on ARC-2 (See Table~\ref{tab:main_results}).

\noindent
\textbf{Total Loss.} The overall objective combines the two terms:
\begin{equation}
    \mathcal{L}_{\text{total}} = \mathcal{L}_{\text{task}} + \lambda \mathcal{L}_{\text{align}},
\end{equation}
where $\lambda$ balances the two losses. We set $\lambda=0.1$; as shown in Sec.~\ref{sec:lambda}, performance is stable for $\lambda\in[0.05,0.15]$, with over-regularization at $\lambda=0.2$.

\begin{figure*}[t]
    \centering
    \includegraphics[width=1.0\linewidth]{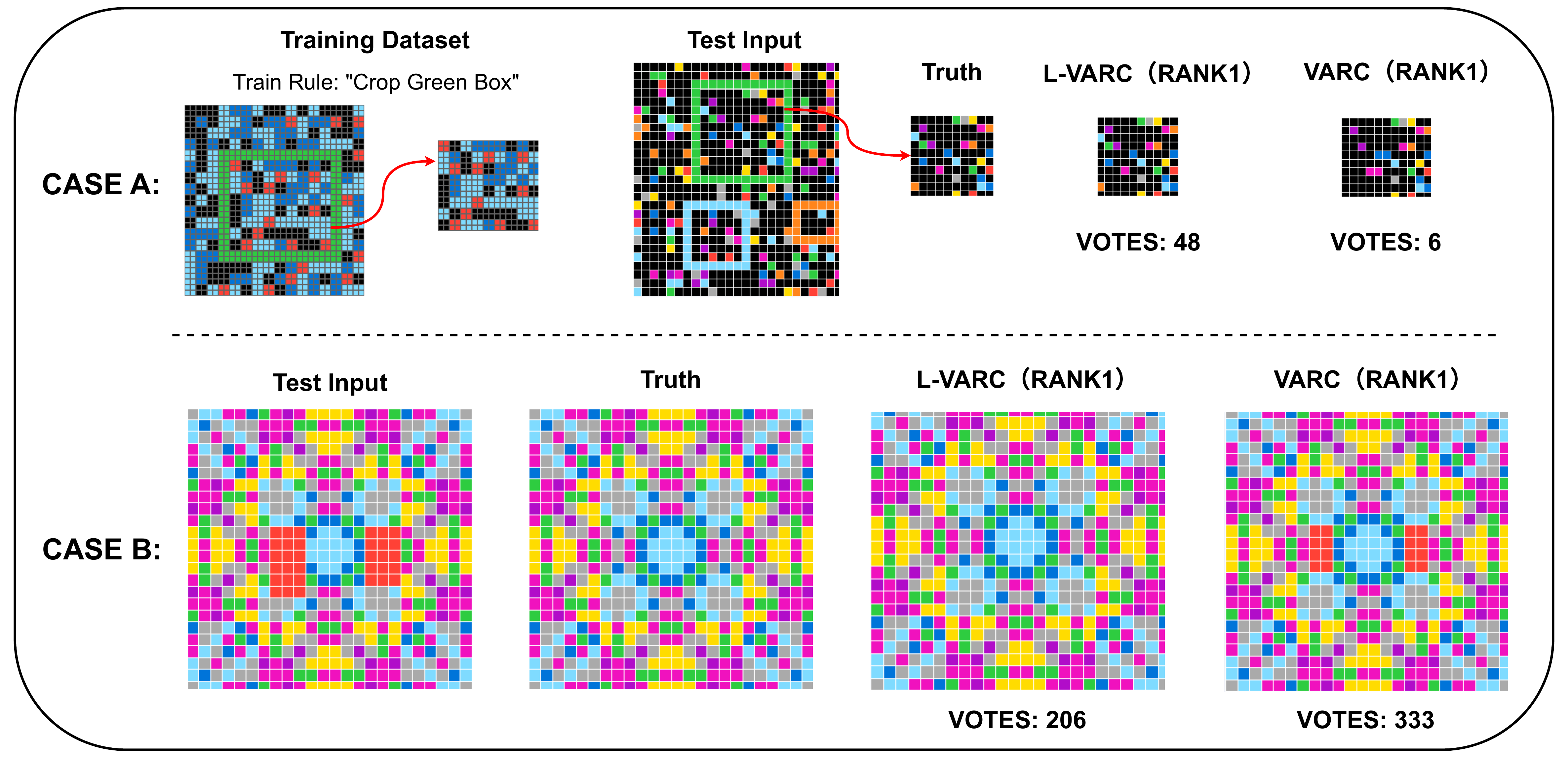} 
    \vspace{1.0em}
    \caption{\textbf{Qualitative comparison on unseen test tasks.} \textbf{Case A (Logic Transfer):} The training task demonstrates a ``crop green object" rule. In the test input, despite the presence of distractors (blue/orange boxes), L-VARC correctly generalizes the rule (crop largest object) to crop the target green box, showing significantly higher voting confidence (48) than the baseline (6). \textbf{Case B (Complex Filling):} L-VARC demonstrates superior fine-grained localization, completely filling the target regions, whereas the baseline fails to cover all pixels (missing spots).}
    \label{fig:qualitative}
\end{figure*}

\subsection{Semantic Compression in Practice}
\label{sec:scm_example}

Fig.~\ref{fig:scm_example} shows a concrete example of the proposed semantic compression module. The raw LARC description for a task often contains multiple sentences, contradictory annotations, or irrelevant perceptual details. For instance:

\noindent
\textit{Raw:} ``create a black 9$\times$9 grid. Copy the pattern of the input grid onto the output grid. However, the number of copies on the output grid is determined by the number of colored squares on the input grid ...''

\noindent
\textit{Refined by DeepSeek-V3:} ``Task: Copy input pattern N times onto 9$\times$9 grid, N = colored squares count, positions match input colored squares | Size: 9$\times$9.''

The refined text (averaging 42 tokens) fits easily within CLIP's 77‑token limit, whereas the raw version (averaging 154 tokens) would be truncated by 50\% of tasks. Such an arbitrary truncation leads to serious loss of semantic information. Note that our compression step is performed offline once and the resulting refined text dataset has been released with our code at https://github.com/GZHU-DVL/L-VARC.

\subsection{Inference with Test-Time Training}
During inference, the ground-truth transformation rule is unknown. Therefore, we discard the entire language branch (DeepSeek, CLIP, and Projector). The model relies solely on the visual backbone. Following the VARC protocol, we apply Test-Time Training for inference target. For each test task, we freeze the main encoder and fine-tune the task tokens and lightweight decoder on the support set $S$ for a few iterations. This allows the semantically-enriched visual encoder to adapt to the specific rule of the unseen task.

\begin{center}
    \includegraphics[width=0.9\linewidth]{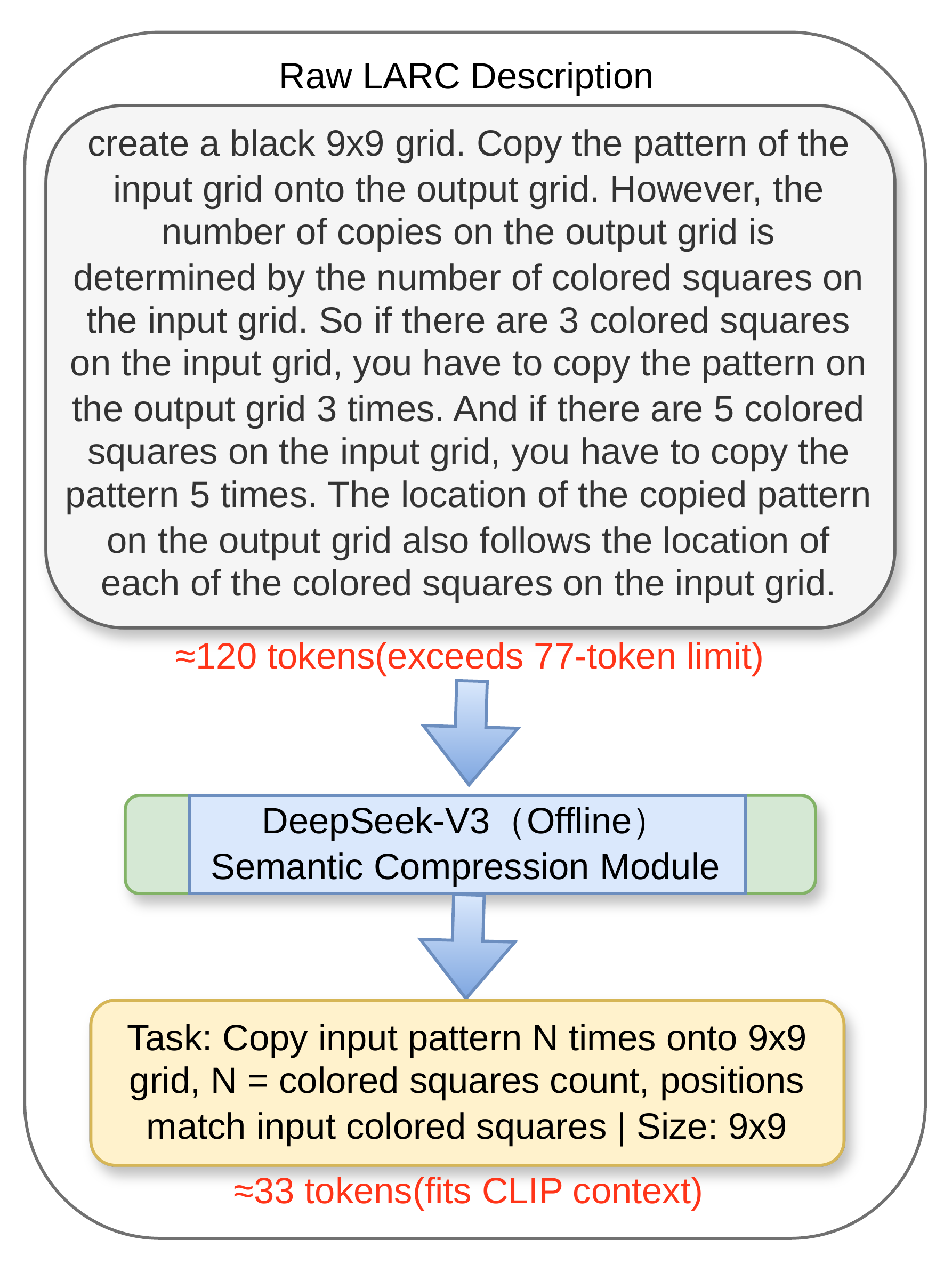}
    \captionof{figure}{Illustration of the semantic compression module. Raw human descriptions (e.g., 120 tokens) exceed CLIP's 77-token limit; DeepSeek-V3 distills them into concise rule statements (e.g., 33 tokens) that fit the context length, avoiding truncation.}
    \label{fig:scm_example}
\end{center}

\section{Experimental Results}
\label{sec:experiments}

\subsection{Experimental Setup}
\textbf{Datasets.} We evaluate our method on the standard ARC evaluation set (ARC-1) \cite{chollet2019measure}, which contains 400 unseen tasks. To test generalization on out-of-distribution tasks, we also evaluate our L-VARC on the challenging ARC-2 subset.
\textbf{Metrics.} We report \textbf{PASS@1} (success rate with 1 attempt) and \textbf{ORACLE} (success rate if the best of $k$ predictions is selected). Additionally, we monitor the \textbf{Eval Accuracy} (pixel-wise accuracy on validation set) during training to analyze learning stability.

\noindent
\textbf{Offline LLM cost.} As illustrated in Sec.~\ref{sec:scm_example}, the semantic compression step uses DeepSeek-V3 \cite{liu2024deepseek} offline one time.

\subsection{Quantitative Comparison}
\textbf{Comparison with State-of-the-Art.}
First, we position our proposed L-VARC against existing Neural and Neuro-Symbolic approaches. As shown in Table \ref{tab:sota}, early methods like HRM \cite{wang2025hierarchical} and TRM \cite{jolicoeur2025less} struggle to surpass 45\% in PASS@2. While the SOTA VARC significantly advanced this field, our L-VARC further pushes the boundary and performs the best on the strict PASS@1 metric among neural approaches. Note that our L-VARC remains the same model size as the baseline VARC, with 1M more parameters than VARC only during training.

\begin{table}[h]
    \centering
    \caption{Performance comparison (\%) with SOTA methods on ARC-1. Results for HRM and TRM are reported from their original papers. The best results are marked in bold.}
    \vspace{1.5em}
    \label{tab:sota}
    \setlength{\tabcolsep}{5pt}
    \begin{tabular}{l c c c}
        \toprule
        \textbf{Method} & \textbf{Param} & \textbf{PASS@1} & \textbf{PASS@2} \\
        \midrule
        HRM \cite{wang2025hierarchical} & 27M & - & 41.25 \\
        TRM \cite{jolicoeur2025less} & 7M & - & 43.50 \\
        VARC \cite{hu2025arc} & 18M & 49.75 & \textbf{55.63} \\
        \midrule
        \textbf{L-VARC (Ours)} & 18M & \textbf{50.62} & 55.37 \\
        \bottomrule
    \end{tabular}
\end{table}

\begin{table}[h]
    \centering
    \caption{Performance of recent large language models on ARC-1 and ARC-2 (solve rates, as reported in respective sources). Our L-VARC (18M params) is shown for reference.}
    \vspace{1.5em}
    \label{tab:llm_comparison}
    \begin{tabular}{l c c c}
        \toprule
        \textbf{Method} & \textbf{Params} & \textbf{ARC-1 (\%)} & \textbf{ARC-2 (\%)} \\
        \midrule
        DeepSeek R1 & 671B & 15.8 & 1.3 \\
        Claude 3.7 & N/A & 21.2 & 0.9 \\
        o3-mini-high & N/A & 34.5 & 3.0 \\
        GPT-5 & N/A & 44.0 & 1.9 \\
        Grok-4-thinking & 1.7T & 66.7 & 16.0 \\
        Bespoke (Grok-4) & 1.7T & 79.6 & 29.4 \\
        \midrule
        \textbf{L-VARC (Ours)} & \textbf{18M} & \textbf{55.37} & \textbf{7.50} \\
        \bottomrule
    \end{tabular}
\end{table}

\noindent
\textbf{Detailed Comparison with Baseline.}
We provide a comprehensive comparison with the strongest baseline, VARC, on both ARC-1 and ARC-2 in Table \ref{tab:main_results}. As shown in Table~\ref{tab:llm_comparison}, although trillion-parameter LLMs, such as Bespoke (Grok-4), achieve higher absolute accuracy, they require massive computational resources and do not adhere to the ``scratch'' training constraint. In contrast, our L-VARC obtains competitive results with only 18M parameters.
L-VARC outperforms the baseline on the primary ARC-1 benchmark, achieving a \textbf{PASS@1 of 50.62\%}. The improvement in \textbf{ORACLE} (64.50\% vs 63.75\%) indicates that our language-aligned encoder effectively retrieves correct hypotheses within top candidates.
Crucially, on the hard \textbf{ARC-2} dataset, our method improves the reproduced baseline from 6.25\% to \textbf{6.67\%}. This suggests that the abstract rules learned via language alignment transfer better to new, out-of-distribution tasks than purely visual features.

\begin{table}[h]
    \centering
    \caption{Detailed performance (\%) on ARC-1 and ARC-2. * indicates results reproduced in our experiments for fair comparison. The best results are marked in bold.}
    \vspace{1.5em}
    \label{tab:main_results}
    \setlength{\tabcolsep}{3.5pt}
    \begin{tabular}{l | c c c | c c c}
        \toprule
        \multirow{2}{*}{\textbf{Method}} & \multicolumn{3}{c|}{\textbf{ARC-1 (Standard)}} & \multicolumn{3}{c}{\textbf{ARC-2 (Hard)}} \\
         & P@1 & P@2 & Oracle & P@1 & P@2 & Oracle \\
        \midrule
        VARC \cite{hu2025arc} & 49.75* & \textbf{55.63*} & 63.75* & 6.25* & 7.08* & \textbf{8.75*} \\
        \textbf{L-VARC} & \textbf{50.62} & 55.37 & \textbf{64.50} & \textbf{6.67} & \textbf{7.50} & 8.61 \\
        \midrule
        \textit{Gain} & \textit{+0.87} & \textit{-0.26} & \textit{+0.75} & \textit{+0.42} & \textit{+0.42} & \textit{-0.14} \\
        \bottomrule
    \end{tabular}
    \vspace{-1em}
\end{table}

\subsection{Qualitative Analysis}
To provide deeper insights, we visualize representative success cases in Fig. \ref{fig:qualitative} and analyze the model behavior from the perspectives of logic robustness and pixel-level localization. The ``Votes" metric indicates the frequency of the top-1 prediction appearing across 510 test-time training samples.

\noindent
\textbf{Robustness to Distractors and Confidence (Case A).}  As shown in Case A of Fig. \ref{fig:qualitative}, the model must learn to ignore distractors (new colors appearing in the test set) and strictly follow the ``crop green object" rule derived from training. Our L-VARC successfully transfers this abstract logic to the test case (crop the largest object). Notably, the voting statistics reveal a sharp contrast in model uncertainty: L-VARC converges to the correct solution with \textbf{48 votes}, whereas the baseline's top prediction is incorrect and supported by only \textbf{6 votes}. This indicates that without language guidance, the baseline struggles to find a consistent hypothesis, while our L-VARC forms a much sharper and more confident decision boundary.

\noindent
\textbf{Fine-grained Localization (Case B).} For tasks requiring dense pixel manipulation, such as the complex filling task in Case B, we observe that L-VARC exhibits more precise spatial attention. The baseline, despite converging to a specific pattern (high votes), falls into a suboptimal solution with incomplete outputs  (``missing spots"). In contrast, L-VARC correctly identifies and fills all target regions. We attribute this to the Cross-Attention Projector, which forces the visual encoder to align with semantic concepts like ``fill" or ``complete" during the training phase, preventing the model from ignoring hard-to-predict edge pixels.

\subsection{Hyperparameter Sensitivity}
\label{sec:lambda}
The total loss includes a weight $\lambda$ for the alignment term $\mathcal{L}_{align}$. We set $\lambda=0.1$ in all main experiments. Fig.~\ref{fig:lambda_sensitivity} shows the PASS@1 on ARC‑1 ($\lambda \in \{0.05, 0.1, 0.15, 0.2\}$). Performance remains stable within $\pm0.2\%$ for $\lambda \in [0.05, 0.15]$, with a drop of $0.5\%$ at $\lambda=0.2$ due to over‑regularization. The recommended range $0.1\pm0.05$ is reliable for reproduction.

\begin{center}
    \includegraphics[trim={0cm 0cm 0cm 0cm},clip,width=1\linewidth]{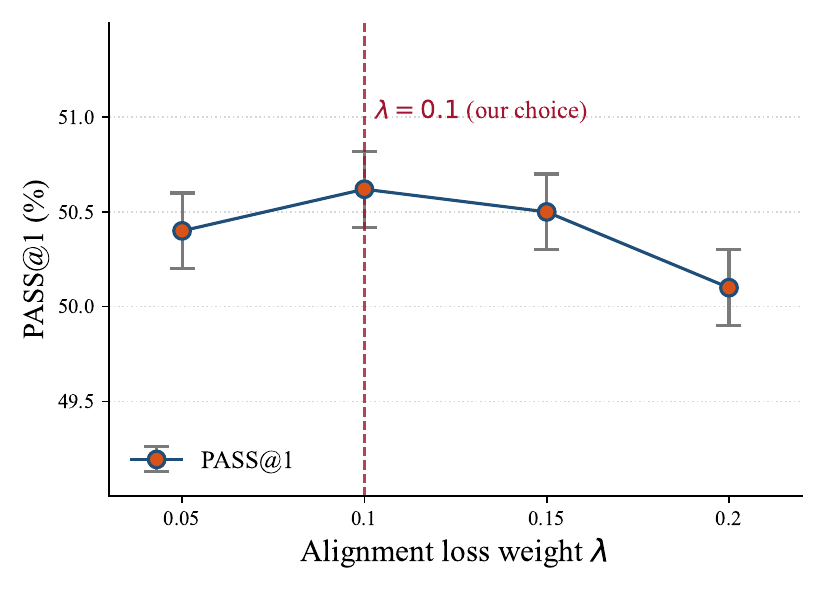} 
    \captionof{figure}{Sensitivity to the alignment loss weight $\lambda$. Performance remains stable within $\pm0.2\%$ for $\lambda\in[0.05,0.15]$, with a drop at $\lambda=0.2$ due to over-regularization. The recommended range $0.1\pm0.05$ is safe for reproduction.}
    \label{fig:lambda_sensitivity}
\end{center}

\subsection{Ablation Study}
To confirm the contribution of the proposed two components, we compare the cross-attention projector (CAP) against a global average pooling (GAP) baseline, both utilizing the proposed semantic compression module (SCM). As shown in Table \ref{tab:ablation} and Fig. \ref{fig:curve}, our SCM improves over the vision-only baseline VARC (0.7764 vs. 0.7668). This is due to the fact our SCM can effectively condense the raw text description to concise semantic embeddings. Besides, the L-VARC (i.e., VARC+SCM+CAP) achieves the highest accuracy of 0.7812. This performance gain confirms that our CAP enables a more robust mapping between visual patterns and semantic rules, which further validates that ARC tasks require dynamically attending to specific, task-relevant local regions (e.g., discrete objects) rather than global feature averaging, like GAP.

\begin{center}
    \includegraphics[trim={0cm 0cm 0cm 0cm},clip,width=0.8\linewidth]{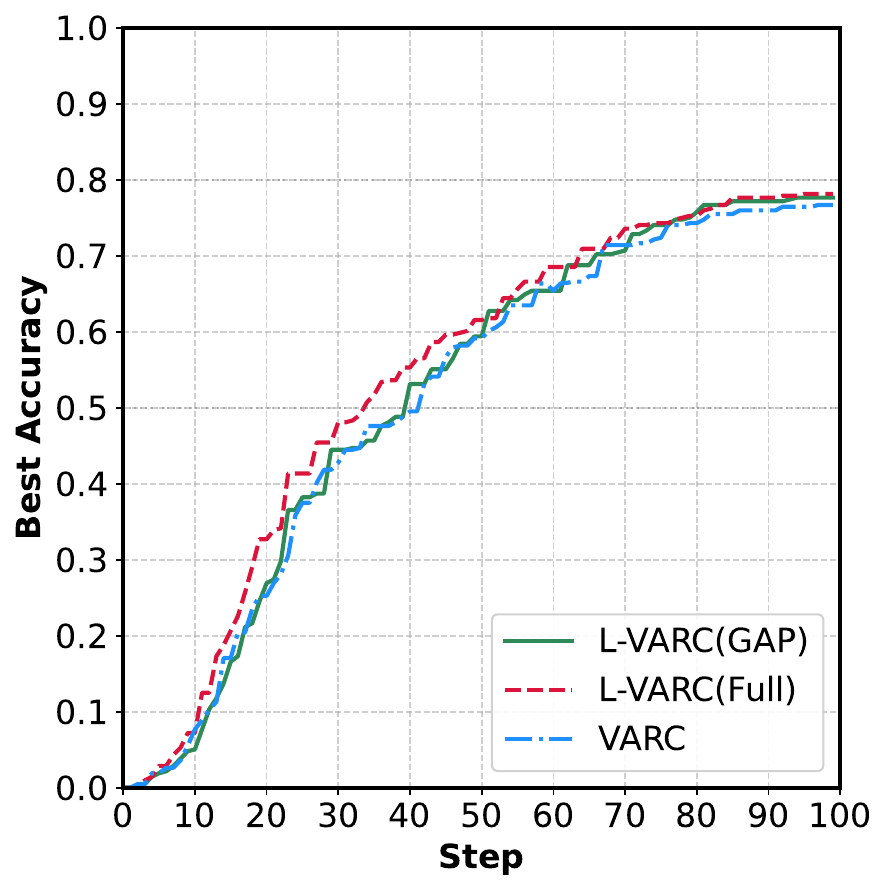} 
    \captionof{figure}{Training Dynamics. L-VARC (CAP) demonstrates superior convergence and peak accuracy compared to the GAP variant and vision-only baseline.}
    \label{fig:curve}
\end{center}

\begin{table}[h]
    \centering
    \caption{Ablation results on the proposed components.}
    \label{tab:ablation}
    \setlength{\tabcolsep}{8pt}
    \begin{tabular}{lcc}
        \toprule
        \textbf{Method Setting} & \textbf{Projector} & \textbf{Best Eval Acc.} \\
        \midrule
        Baseline \cite{hu2025arc} & N/A & 0.7668 \\
        L-VARC (SCM) & GAP & 0.7764 \\
        \rowcolor[HTML]{EFEFEF} \textbf{L-VARC (Full)} & \textbf{CAP} & \textbf{0.7812} \\
        \bottomrule
    \end{tabular}
\end{table}

\noindent
\textbf{Statistical robustness.} To show statistical robustness, we report mean and standard deviation over 3 runs. As shown in Fig.~\ref{fig:stability}, the per-run PASS@1 values for VARC are 49.46, 49.75, and 49.91 (mean $49.75\pm0.3$) and for L-VARC are 50.42, 50.60, and 50.78 (mean $50.62\pm0.3$). The improvement $+0.87$ exceeds the run variance, confirming that this advantage is not due to random seed luck. The gain corresponds to 3–4 additional tasks solved on the 400‑task ARC‑1 set.

\begin{center}
    \includegraphics[width=1\linewidth]{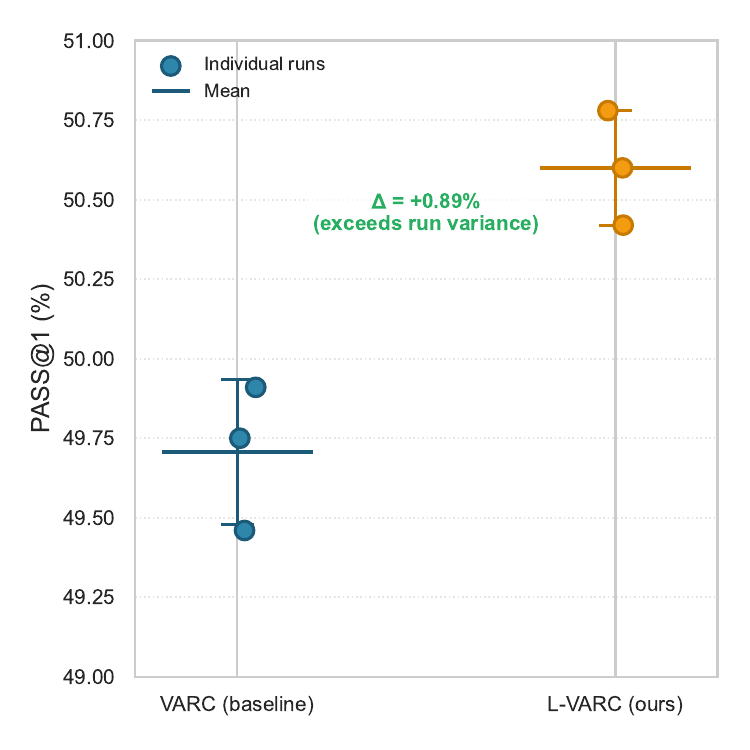}
    \captionof{figure}{Three-run distribution of PASS@1 on ARC-1. Each point represents one run; horizontal lines and error bars indicate mean $\pm$ standard deviation. L-VARC consistently outperforms VARC across all three seeds, with an average gain of $+0.87\%$, which exceeds the within-run variance.}
    \label{fig:stability}
\end{center}

\begin{table}[h]
    \centering
    \caption{Effect of raw (uncompressed) LARC descriptions. Using raw text (truncated to 77 tokens) degrades PASS@1 by 2.0\% absolute, demonstrating that our SCM prevents catastrophic information loss.}
    \label{tab:raw_ablation}
    \begin{tabular}{lc}
        \toprule
        \textbf{Method} & \textbf{PASS@1 (\%)} \\
        \midrule
        VARC (baseline) & 49.75 \\
        + Raw LARC (truncated) & 47.75 (-2.0) \\
        + SCM (compressed) & 50.62 (+0.87) \\
        \bottomrule
    \end{tabular}
\end{table}

\subsection{Per‑Category Generalization Analysis}
\label{sec:percat}
To better understand where language guidance provides the most benefit, we manually categorize a subset of ARC-1 tasks into four groups based on their dominant transformation rule: \textit{Expansion} (e.g., scaling, diffusion), \textit{Transformation} (e.g., color inversion, flipping), \textit{Filling} (e.g., enclosing regions), and \textit{Cropping} (e.g., extracting largest object). As shown in Fig.~\ref{fig:percat}, L-VARC solves more tasks than VARC in three categories. The most notable gain occurs in \textit{Transformation} (+8 tasks), where semantic descriptions help the model capture abstract operations like color inversion. \textit{Filling} also benefits substantially (+6 tasks), likely because language clarifies the concept of “complete filling”. The small decrease on \textit{Expansion} (1 task) is within run variance, and \textit{Cropping} improves modestly. These results support that language guidance is particularly effective for tasks requiring conceptual rule induction rather than low‑level pattern matching.

\begin{center}
    \includegraphics[trim={0cm 0cm 0cm 0cm},clip,width=1\linewidth]{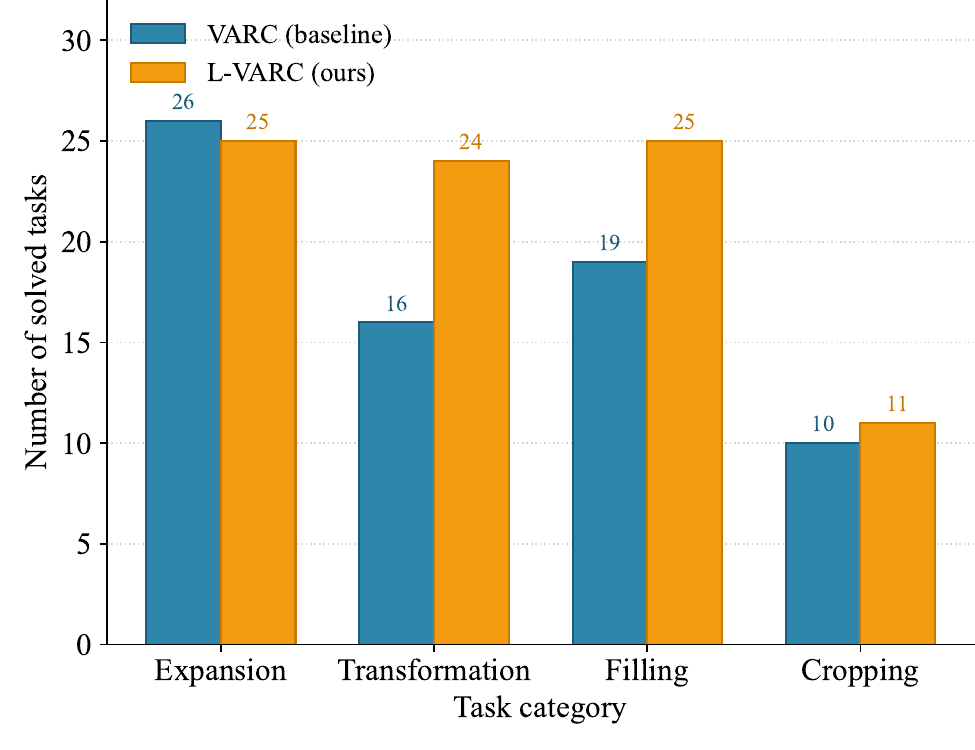} 
    \vspace{-1em}
    \captionof{figure}{Number of successfully solved tasks per category on ARC-1. L-VARC consistently outperforms VARC in three out of four categories, with the largest absolute gain (+8 tasks) on \textit{Transformation} tasks. While \textit{Filling} and \textit{Cropping} improve by 6 and 1 tasks, respectively. These categories are manually defined subsets of ARC-1 where the underlying rule is relatively unambiguous.}
    \label{fig:percat}
\end{center}

\FloatBarrier
\section{Discussion}
\label{sec:discussion}

\subsection{Transfer vs. Memorization}
The observed performance gain ($+0.87\%$ on ARC-1, $+0.42\%$ on ARC-2) raises a fundamental question: does this improvement truly reflect abstraction transfer, or merely an improved inductive bias within the same task distribution? While a definitive causal separation would require controlled experiments on synthetic task distributions---which we leave for future work---several pieces of evidence support the transfer interpretation \cite{wang2024cognition}.

\textbf{ARC's strict evaluation protocol.} The ARC-1 test set is designed to be completely disjoint from the training set in terms of underlying core knowledge priors \cite{chollet2019measure}. Any form of pixel-level memorization would not generalize to tasks governed by entirely new transformation rules. The fact that L-VARC consistently solves 3--4 additional unseen tasks across three independent runs suggests that the model has acquired more generalizable knowledge.

\textbf{Gain on out-of-distribution ARC-2.} The ARC-2 subset is specifically constructed to be more challenging and to require combining multiple core knowledge priors in new ways. The fact that L-VARC also improves on ARC-2 ($+0.42\%$), albeit modestly, indicates that the learned representations capture features that transfer beyond the exact training distribution.

\textbf{Per-category analysis.} As shown in Sec.~\ref{sec:percat}, the largest improvements occur on the \textit{Transformation} and \textit{Filling} tasks, which demand conceptual rule induction. If the gain is purely due to improved inductive bias within a fixed task distribution, we would expect more uniform improvements across categories.

\textbf{Broader context from recent surveys.} A comprehensive living survey of ARC research, covering 82 approaches across three benchmark versions, finds that performance degradation across versions is consistent across all paradigms (program synthesis, neuro-symbolic, and neural approaches), indicating fundamental limitations in compositional generalization \cite{vahdati2026arc}. Notably, while systems now reach 93.0\% on ARC-AGI-1 (e.g., Opus 4.6), performance falls to 68.8\% on ARC-AGI-2 and further to only 13\% on ARC-AGI-3, whereas humans maintain near-perfect accuracy across all versions \cite{vahdati2026arc}. This substantial performance drop highlights that even the most advanced systems---including trillion-scale models---struggle with true compositional reasoning. Our modest but robust improvement on ARC-2 should be understood within this context: the \emph{relative} gain of $+0.42\%$ on a benchmark where most models struggle to reach double-digit accuracy is non-trivial. According to the survey notes \cite{vahdati2026arc}, ARC-v1 has seen extensive exploration over six years, yet only with the release of ARC-v2 in 2025 has the research community begun to systematically address the challenge of out-of-distribution generalization. Our proposed L-VARC can be situated within the broader taxonomy proposed by recent ARC surveys \cite{anonymous2026a}, performing inductive learning via offline training and transductive inference via test-time training.

Nevertheless, we acknowledge limitations in establishing causality. Future work could systematically vary the similarity between training and test task distributions to quantify the memorization vs. transfer tradeoff explicitly, e.g., using the controlled dataset generation framework proposed by recent ARC-AGI-2 studies \cite{de2026arc}.

\subsection{Failure Modes and Remaining Challenges}
Despite the overall improvement, L-VARC still fails on several task types. Through manual inspection, we identify two dominant failure modes:

\begin{itemize}
    \item \textbf{Multi-step arithmetic tasks}: Tasks that involve counting or arithmetic operations over multiple steps (e.g., ``add the number of blue squares, then subtract the number of red squares'') often stump the model. The language descriptions for such tasks are sometimes ambiguous, and the cross-attention projector may attend to irrelevant spatial regions.
    \item \textbf{Recursive / fractal patterns}: Tasks that require generating self-similar patterns (e.g., infinite nesting) are rarely solved. The current ViT backbone with fixed receptive fields struggles to capture long-range dependencies of recursive rules.
\end{itemize}

A recurring theme across the ARC literature is the difficulty of true compositional generalization. The ARC Prize 2025 winners needed hundreds of thousands of synthetic examples to reach only 24\% on ARC-AGI-2, confirming that reasoning remains fundamentally ``knowledge-bound'' rather than purely architecture-bound \cite{vahdati2026arc}. Our failure analysis aligns with this observation: L-VARC performs well when the rule can be described as a simple, localizable transformation (e.g., ``fill the enclosed region''), but breaks down when reasoning requires multi-step state tracking or recursive pattern generation \cite{acosta2025survey}.

Addressing these limitations may require incorporating iterative reasoning modules or dynamic computation graphs that can adapt to the structure of the rule. LoopViT \cite{shu2026loopvit} offers a concrete example: by iterating a weight‑tied hybrid block with a dynamic exit mechanism, it demonstrates that adaptive iterative computation can effectively address recursive pattern generation at a fraction of the parameter cost. One promising direction is to extend test-time training to include iterative hypothesis refinement, where the model generates candidate rules, tests them against the support set, and iteratively refines its predictions. This aligns with the emerging consensus that \emph{test-time adaptation and refinement loops are critical success factors} for ARC solvers \cite{vahdati2026arc}.

Another promising avenue is to incorporate structural priors that explicitly encode symmetry and invariance. The ARC-AGI-2 transformer system, for example, introduces a principled augmentation framework based on group symmetries, grid traversals, and automata perturbations, enforcing invariance to representation changes \cite{de2026arc}. Such symmetry-aware encoding could be naturally integrated into our framework to further improve generalization on tasks where the transformation rule is invariant to rotations, reflections, or color permutations.

\subsection{Computational Cost and Reproducibility}
DeepSeek-V3 is used only \textbf{offline once} to compress the LARC descriptions. The refined text dataset (averaging 36 tokens per task) has been released along with the source code. Therefore, any researcher can reproduce our work without running an LLM. The actual training of L-VARC takes 32 hours on two RTX 4090 GPUs, and test-time inference (including TTT) takes about 4 minutes per task---comparable to the baseline VARC. The entire code and data are publicly available.

We emphasize that our framework operates under the Learning Using Privileged Information (LUPI) paradigm originally introduced by Vapnik \cite{pechyony2010theory}. Recent advances have extended LUPI to incorporate statistical invariants, showing that privileged information can be represented as weak convergence constraints that reduce the hypothesis space without requiring strong distributional assumptions \cite{yan2025learning}. This theoretical perspective justifies our empirical choice to discard the privileged branch at inference time: the privileged information helps the model find a better admissible function during training but is not required for evaluation. Our work thus contributes to a growing body of evidence that LUPI is particularly effective when training data is limited and the privileged information provides complementary structure, as is precisely the case with LARC descriptions for ARC tasks.

\subsection{Broader Impact and Future Directions}
Our work demonstrates that even a small amount of high-quality linguistic supervision can substantially improve visual reasoning. This suggests several promising directions for future research.

\textbf{Iterative and adaptive reasoning.} Beyond static privileged information, future models could interact with an LLM dynamically during test time to refine hypotheses. This direction is inspired by recent advances in multimodal reasoning where large language models serve as meta-reasoners for visual tasks, generating counterfactual samples or structured reasoning traces \cite{song2025codedance, zhang2025llm}. For ARC, an iterative ``hypothesis-propose-verify'' loop could allow the model to generate candidate rules in natural language, test them against the support set, and refine its predictions accordingly.

\textbf{Visual latent reasoning.} Recent work on visual latent reasoning suggests that models can reason through compact visual representations rather than verbose textual chains, achieving substantial efficiency gains \cite{jiang2026univlr}. Extending our framework to support reasoning directly in the visual latent space---where the model generates intermediate visual states rather than linguistic rules---could unlock more efficient and interpretable inference.

\textbf{Symbolic grounding and representation bottlenecks.} Recent studies on abstract visual reasoning have identified representation quality as a key bottleneck. On Bongard-LOGO problems, for instance, vision-language models often fail not because of reasoning deficits but because their visual representations lack sufficient structure to support abstract rule induction \cite{vaishnav2026symbolic}. Our cross-attention projector partially addresses this by dynamically extracting task-relevant features, but more principled approaches to symbolic grounding --- for example, learning disentangled representations that separate objects, relations, and transformations --- remain an open challenge.

\textbf{From vision-language to language-vision synergy.} A complementary direction is to reverse the information flow: instead of using language to guide visual reasoning, future work could use visual reasoning to ground and refine linguistic representations. This is particularly relevant for tasks where the transformation rule is inherently spatial and cannot be easily described in language. 

\textbf{Scalability to larger task distributions.} Our method is currently limited to the LARC-annotated tasks (approximately 400 training tasks). However, the LUPI paradigm can in principle scale to much larger datasets. Future work could leverage LLMs as offline annotators to generate structured descriptions for arbitrary visual tasks, then distill that knowledge into lightweight vision models. This aligns with recent efforts to automatically synthesize abstract images and reasoning instructions using language models and their code capabilities \cite{zhang-etal-2024-multimodal}. As the cost of LLM inference continues to fall---cost per ARC-AGI task dropped 390-fold from o3's \$4,500 to GPT-5.2's \$12 in just one year \cite{vahdati2026arc}---large-scale offline annotation becomes increasingly feasible.

\textbf{Beyond ARC-AGI: generalization to novel benchmarks.} Finally, while our evaluation focuses on ARC-1 and ARC-2, future work should assess generalization to other abstract reasoning benchmarks such as ConceptARC, Bongard-LOGO, or Raven's Progressive Matrices. Such cross-benchmark evaluation would provide a more comprehensive assessment of whether the language-guided representations learned by L-VARC capture truly generalizable reasoning skills or are specific to the ARC task distribution. Beyond the ARC benchmark, the core idea of L-VARC—leveraging linguistic descriptions as privileged information to guide visual abstraction—naturally extends to medical domains where visual inspection and semantic reasoning are inherently intertwined. In Traditional Chinese Medicine (TCM), inspection diagnosis relies on observing patients' physical signs (e.g., gait abnormalities in stroke, resting tremor in Parkinson's disease) and tongue manifestations (e.g., coating color, body shape) to infer underlying syndrome patterns. These diagnostic processes parallel ARC tasks: the clinician must extract abstract rules from visual observations, often guided by textual medical knowledge. A promising future direction is to reformulate TCM inspection as a language-guided visual reasoning task under the LUPI paradigm, where classical TCM texts serve as privileged semantic priors to regularize deep visual models. Furthermore, the holistic philosophy of TCM resonates with our framework's design—integrating linguistic and visual modalities as a unified whole rather than isolated streams. By distilling TCM diagnostic principles into structured language prompts, one could train lightweight vision models that mimic expert-level inspection reasoning at low computational cost, potentially enabling accessible, interpretable AI-assisted diagnosis in resource-limited clinical settings.

\FloatBarrier
\section{Conclusion}
\label{sec:conclusion}
In this work, we have presented L-VARC, a novel framework that injects semantic reasoning into vision-based ARC solvers through the Learning Using Privileged Information (LUPI) paradigm. By aligning visual representations with refined and structured language rules during training, our model learns more robust and generalized feature abstractions while maintaining a lightweight inference footprint of only 18 million parameters.

Our key findings are threefold. First, we demonstrated that raw LARC descriptions, when directly encoded by CLIP, suffer from severe truncation (over 60\% of tasks exceed the 77-token limit), causing a 2.0\% absolute drop in PASS@1. The proposed Semantic Compression Module (SCM) using DeepSeek-V3 effectively distills verbose descriptions into concise rule statements (average 36 tokens), eliminating truncation and recovering the lost performance. Second, the Cross-Attention Projector (CAP) dynamically extracts task-relevant visual features and aligns them with semantic embeddings, outperforming simple global average pooling by 0.48\% in evaluation accuracy. Third, language guidance consistently improves generalization: on ARC-1, L-VARC achieves a +0.87\% PASS@1 gain over the strong VARC baseline (49.75\% → 50.62\%), corresponding to 3–4 additional tasks solved across three independent runs. On the more challenging out-of-distribution ARC-2 benchmark, our method obtains a +0.42\% gain (6.25\% → 6.67\%), with per-category analysis showing the largest improvements on \textit{Transformation} (+8 tasks) and \textit{Filling} (+6 tasks) — categories that demand conceptual rule induction rather than low-level pattern matching.



\printcredits

\end{document}